\documentclass[oneside]{llncs}

\usepackage[T1]{fontenc}

\usepackage[fleqn]{mathtools} 
\usepackage{amsfonts}
\usepackage{etoolbox}

\usepackage{newtxtext,newtxmath}
\usepackage[scaled=0.93,proportional]{zlmtt}

\usepackage{graphicx}
\usepackage{hyperref}
\usepackage{color}

\usepackage[acronym]{glossaries}
\usepackage[useregional]{datetime2}
\usepackage{hyperref}
\usepackage{xspace}
\usepackage{enumitem}
\usepackage{tikz, subcaption}
\usetikzlibrary{arrows.meta}
\usepackage{bm}
\usepackage{geometry, pdflscape, ltablex, ragged2e, multirow,booktabs}
\usepackage{tikz}
\usepackage{booktabs}
\usepackage{multirow}
\usetikzlibrary{shapes.geometric, arrows.meta, positioning, calc}
\usepackage{url}
\usepackage{pgfplots} 
\usepackage{pgfplotstable}
\pgfplotsset{width=15.5cm, height=6.5cm, compat=1.18}  

\usepackage{algorithm}
\usepackage{algpseudocode}
\usepackage{pseudo}
\usepackage{placeins}

\usepackage{appendix}

\usepackage{tabularx}
\usepackage{float}
\usepackage{parskip}
\graphicspath{{figures}}

\usepackage{graphicx}

\definecolor{DodgerUniformBlue}{rgb}{0.0,0.353,0.612}
\newcommand{\define}[1]{\emph{\textcolor{DodgerUniformBlue}{#1}}}

\definecolor{paleyellow}{HTML}{FFEC7F}

\long\def\comment[#1]#2{\medskip\par\noindent\colorbox{paleyellow}{\llap{\footnotesize #1:\quad}%
    \parbox[t]{\textwidth}{\setlength{\parskip}{1ex plus 0.2ex minus 0.2ex}#2}}}

\def\void{}
\newcommand{\mcomment}[2][\void]{\marginpar{\raggedright\tiny\ifx\void#1\else\textbf{#1:\\}\fi#2}}

\newcolumntype{L}{>{$}l<{$}}%
\newcolumntype{C}{>{$}c<{$}}%
\newcolumntype{R}{>{$}r<{$}}%
\newcolumntype{M}{@{}p{\mathindent}@{}}%
\newcolumntype{t}{>{\mbox\bgroup}l<{\egroup}}%
\newcolumntype{e}{r@{\;}l}%
\newcolumntype{E}{>{$}r<{$}@{$\;$}>{$}l<{$}}%

\renewcommand{\emph}[1]{{\rmfamily\itshape #1}}

\makeatletter
\renewcommand*\cleardoublepage{\clearpage}
\makeatother

\makeatletter
\patchcmd{\thebibliography}{\clearpage}{}{}{}
\makeatother

\makeatletter
\patchcmd{\thebibliography}{\cleardoublepage}{}{}{}
\makeatother

\begin{document}
\title{Large Language Models Imitate Logical Reasoning, but at what Cost?}
\titlerunning{Large Language Models' Reasoning Stalls}
%
\author{Lachlan McGinness\inst{1,2}\orcidID{0000-0002-3231-4827},   Peter Baumgartner\inst{2,1}\orcidID{0000-0002-6559-9654}} 
%
\authorrunning{L. McGinness, P. Baumgartner}
%
\institute{Australian National University, 
\and
Data61|CSIRO, 
}
\maketitle              
\begin{abstract}
We present a longitudinal study which evaluates the reasoning capability of frontier Large Language Models over an eighteen month period. We measured the accuracy of three leading models from December 2023, September 2024 and June 2025 on true or false questions from the PrOntoQA dataset and their faithfulness to reasoning strategies provided through in-context learning. The improvement in performance from 2023 to 2024 can be attributed to hidden Chain of Thought prompting. The introduction of thinking models allowed for significant improvement in model performance between 2024 and 2025. 

We then present a neuro-symbolic architecture which uses LLMs of less than 15 billion parameters to translate the problems into a standardised form. We then parse the standardised forms of the problems into a program to be solved by Z3, an SMT solver, to determine the satisfiability of the query. We report the number of prompt and completion tokens as well as the computational cost in FLOPs for open source models. The neuro-symbolic approach significantly reduces the computational cost while maintaining near perfect performance.  The common approximation that the number of inference FLOPs is double the product of the active parameters and total tokens was accurate within 10\% for all experiments. 

This work has been accepted for publication in the proceedings of AJCAI2025.

\end{abstract}

\section{Introduction}
\label{sec:Introduction}

Transformer-based Large Language Models (LLMs) have been shown to be very good at pattern recognition, next token prediction and generating plausible-sounding explanations. Whether or not such models are capable of intelligent reasoning is a topic that is hotly debated with some arguing that they have almost no reasoning capabilities \cite{Shojaee2025Illusion}, others stating that LLMs will be the basis of an artificial general intelligence \cite{XuPoo2023Large} and still others debating that they can play a useful role in reasoning processes \cite{Kambhampati2024LLMModulo}.

Although there is controversy around the reasoning capabilities of LLMs, strict deductive reasoning is a task (among others) that is rigorously achieved by 
symbolic AI approaches such as Satisfiability Modulo Theory (SMT) solvers. Symbolic AIs do not have the flexibility to accept natural language inputs,
and require their input in formal language, such as a dialect of the popular TPTP family~\cite{sutcliffe_tptp_2017}. In 2022 Saparov released PrOntoQA, a benchmark which requires both the flexibility to understand natural language input as well as the ability to rigorously perform deductive reasoning. 
In this paper, we study PrOntoQA problems from the viewpoint of reasoning with LLMs and in combination with symbolic systems.

We present an 18 month study where frontier LLMs from December 2023, September 2024 and June 2025 are tested on the PrOntoQA benchmark. Srivastava notes that LLM benchmarks are often obsolete within two years after publication \cite{Srivastava2023Beyond} because models have reached near-perfect performance, so we consider this to be a longitudinal study for this benchmark. In addition to accuracy, we record prompt and completion tokens as a measure of computational expense. We use in-context learning to provide models with different strategies for solving the problems. We examine the model responses to determine if they contain all steps required for reasoning and are faithful to the requested reasoning strategy when solving the PrOntoQA problems. 

We present a neuro-symbolic framework which uses an LLM to translate the problems to a standardised form to be solved by Z3 \cite{Z3}, an SMT solver. We show that this framework can be used to answer the questions in the dataset with high accuracy at a fraction of the computational cost. 

Our contributions are:
\begin{itemize}
    \item A survey of frontier models at three points during an eighteen month period with an analysis of accuracy, computational expense and reasoning correctness for different prompting techniques.
    \item Demonstration of a neuro-symbolic framework to solve natural language problems which require automated reasoning. 
\end{itemize}

This paper is organised as follows:
In Section \ref{sec:background} we introduce the Large Language Models, the benchmark and automated reasoners. In Section \ref{sec:methods} we introduce our methods for determining correctness, faithfulness to reasoning and our neuro-symbolic framework. In Sections \ref{sec:results} and \ref{sec:discussion} we show the findings from our experiments and discuss the implications. 

\section{Background}
\label{sec:background}

\subsection{Large Language Model Reasoning}
Our study is related, but also different, to the many other theoretical and empirical studies on the topic of measuring and improving LLM reasoning \cite{Agarwal2024Many,Dziri2023Faith,Gao2024Retrieval,Huang2023Reasoning,Kaplan2020Scaling,liu_logical_2025,liu_efficient_2025,Pfau2024Let,Tammet2024,Valmeekam2023Planning}. 
The recent study in~\cite{Shojaee2025Illusion}, for instance, works with logical puzzles. Like PrOntoQA \cite{Saparov2023Language}, they test core deductive reasoning capabilities. The problem difficulty can be scaled by a parameter to make problem instances harder. 
They observe accuracy breakdowns after complexity thresholds are exceeded. 

We do not challenge LLMs to their breakdown limits, but focus on 
(1) resource consumption, (2) the ability of LLMs to follow given instructions for enhancing reasoning accuracy, and (3) soundness and faithfulness of their provided reasoning processes.      
To do this we use the PrOntoQA benchmark which was published in 2023. This allows for a detailed assessment of the LLMs' reasoning process, not just their accuracy in completing the task \cite{Saparov2023Language}. 
PrOntoQA requires models to use deductive reasoning to answer a true or false query.  PrOntoQA is not a static benchmark; it generates problems on demand with different numbers of required steps of reasoning and distractors. This decreases the chance of contamination, where the benchmark is contained in the LLM's traning data.

Two approaches to enhance LLMs' reasoning are what we call \emph{shallow} vs. \emph{deep} embedding of reasoning. Shallow embedding means to instruct LLMs with concrete reasoning strategies to be used by either fine-tuning or in-context learning. In deep embedding, LLMs translate the given reasoning problems into a formal logical syntax so that automated reasoning systems can be used. 

The recent paper~\cite{Morishita2024Enhancing} investigates, in this terminology, a shallow embedding approach. They train LLMs with standard reasoning \emph{patterns}, such as modus ponens, contraposition, syllogism, and logical operator elimination. They fine-tune the LLM on examples of correct and incorrect reasoning patterns. In contrast, our shallow approach uses in-context learning to provide the LLM with reasoning \emph{strategies}.  

\begin{figure}[h!]
    \centering
    \scalebox{0.84}{
    \begin{minipage}[b]{0.45\linewidth}
    \begin{tikzpicture}[node distance=1cm and 1.5cm,
    box/.style={rounded corners, draw, align=center, thick, fill=#1!20},
    arrow/.style={-{Stealth[length=3mm]}, thick},
    arrowlabel/.style={font=\scriptsize, fill=none, inner sep=1pt}]

    \node[box=green] (query) {True or false: Whiskers\\ is an animal.};
    \node[box=orange, below=of query] (rule1) {Every mammal is an animal.};
    \node[box=orange, below=of rule1] (rule2) {Each cat is a mammal.};
    \node[box=red, below=of rule2] (fact) {Whiskers is a cat.};
    \node (reasoningmethod) at ($(query.north) + (0,0.5cm)$) {Bottom Up Reasoning};
    
    \draw[arrow] (rule1) -- (query) node[arrowlabel, pos=0.35, right, align=left, text width=2.2cm] {Therefore Whiskers is an animal};
    \draw[arrow] (rule2) -- (rule1) node[arrowlabel, pos=0.35, right, align=left, text width=2.2cm] {Therefore Whiskers is a mammal};
    \draw[arrow] (fact) -- (rule2) node[arrowlabel, pos=0.5, above, align=center, text width=2.2cm] {};

\end{tikzpicture}
\end{minipage}
\hspace{-2.5cm}
\begin{minipage}[b]{0.45\linewidth}
    \begin{tikzpicture}[node distance=1cm and 1.5cm,
    box/.style={rounded corners, draw, align=center, thick, fill=#1!20},
    arrow/.style={-{Stealth[length=3mm]}, thick},
    arrowlabel/.style={font=\scriptsize, fill=none, inner sep=1pt}]

    \node[box=green] (query) {True or false: Whiskers\\ is an animal.};
    \node[box=orange, below=of query] (rule1) {Every mammal is an animal.};
    \node[box=orange, below=of rule1] (rule2) {Each cat is a mammal.};
    \node[box=red, below=of rule2] (fact) {Whiskers is a cat.};
    \node (reasoningmethod) at ($(query.north) + (0,0.5cm)$) {Top Down Reasoning};
        
    \draw[arrow] (query) -- (rule1) node[arrowlabel, pos=0.5, above, align=left, text width=2.2cm] {};
    \draw[arrow] (rule1) -- (rule2) node[arrowlabel, pos=0.35, right, align=left, text width=2.5cm] {True or False: Whiskers is a mammal};
    \draw[arrow] (rule2) -- (fact) node[arrowlabel, pos=0.35, right, align=left, text width=2.5cm] {True or False: Whiskers is a cat};

\end{tikzpicture}
\end{minipage}
\hspace{-2cm}

    \begin{minipage}[b]{0.9\linewidth}
    \begin{tikzpicture}[node distance=1cm and 2.0cm,
    box/.style={rounded corners, draw, align=center, thick, fill=#1!20},
    arrow/.style={-{Stealth[length=3mm]}, thick},
    arrowlabel/.style={font=\scriptsize, fill=none, inner sep=1pt}]

    \node[box=green] (query) {True or false: Whiskers\\ is an animal.};
    \node[box=orange, below=of query] (rule1) {Every mammal is an animal.};
    \node[box=orange, below=of rule1] (rule2) {Each cat is a mammal.};
    \node[box=red, below=of rule2] (fact) {Whiskers is a cat.};
    \node (reasoningmethod) at ($(query.north) + (0,0.5cm)$) {Magic Set Reasoning};
    \node[box=red, right=of fact] (factir2) {Rex is a dog.};
    \node[box=orange, above=of factir2] (rule3) {Dogs are mammals.};

    \draw[arrow] (rule1) -- (query) node[arrowlabel, pos=0.35, right, align=left, text width=2.2cm] {Therefore Whiskers\\ is an animal};
    \draw[arrow] (rule2) -- (rule1) node[arrowlabel, pos=0.35, right, align=left, text width=2.2cm] {Therefore Whiskers\\ is a mammal};
    \draw[arrow] (fact) -- (rule2) node[arrowlabel, pos=0.5, above, align=center, text width=2.2cm] {};
    
    \draw[blue, dashed, thick, rounded corners] ($(fact.south west)+(-1.2,5.4)$) rectangle ($(fact.north east)+(1,-0.6)$);
    \node[align=left, font=\scriptsize, blue, anchor=north] at ($(fact.south west)+(5.90,5)$) (caption) {First the model uses approximative\\ top down reasoning to identify\\ relevant facts and rules. The remaining \\ facts and rules can then be ignored. };
    \node[align=left, font=\scriptsize, black, anchor=north] at ($(fact.south west)+(5.6,3.8)$) (caption) {Then the model performs bottom \\up reasoning on the reduced set \\of axioms.};
    
    \draw[-{Stealth[length=3mm]}, blue, thick] (2.55,0) -- (2.55,-5);

    \coordinate (cross2center) at (4.2,-4.6); 
    \coordinate (cross3center) at (4.2,-3.1);

    
    \draw[blue, thick] ($(cross2center)+(-1,-0.5)$) -- ($(cross2center)+(1,0.5)$); 
    \draw[blue, thick] ($(cross2center)+(-1,0.5)$) -- ($(cross2center)+(1,-0.5)$);
    
    \draw[blue, thick] ($(cross3center)+(-1,-0.5)$) -- ($(cross3center)+(1,0.5)$); 
    \draw[blue, thick] ($(cross3center)+(-1,0.5)$) -- ($(cross3center)+(1,-0.5)$);
\end{tikzpicture}
\end{minipage}}
    \caption{Illustration of reasoning strategies, modified from \cite{McGinness2024Steamroller}. Facts are shown in red, rules in yellow and queries in green. For simplicity, few or no distractors are shown in the examples. Magic set transformation is a basic ATP search space pruning method.}
    \label{fig:reasoning}
\end{figure}
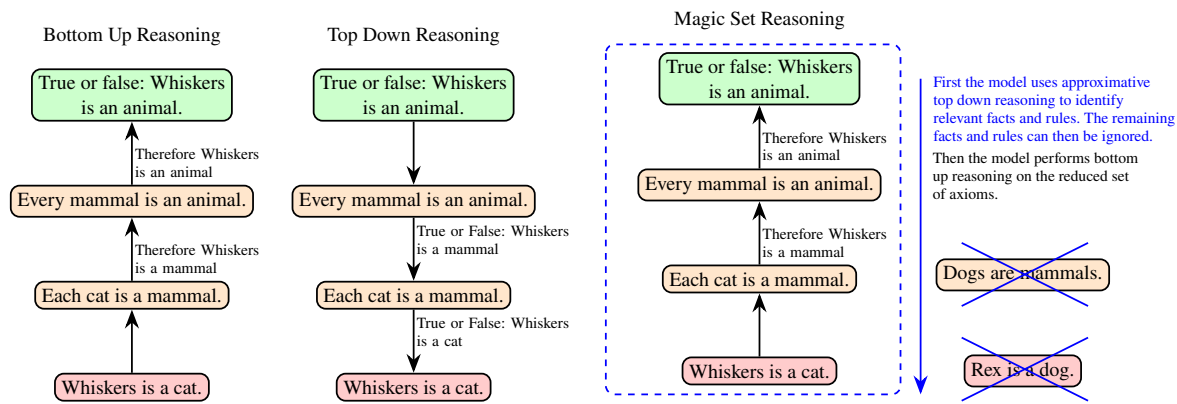

In-context learning or `prompt engineering' provides specific instructions to LLMs to improve their performance on a given task. Chen et al. and Liu et al. provide historical overviews of prompt engineering techniques \cite{Chen2023Unleashing,Liu2023Pre-train}. The most widely used prompt engineering technique for reasoning is Chain of Thought (CoT) \cite{Chen2023Unleashing}, where models think a problem through `step by step'. While we are including CoT in our approach, we do not research CoT fundamentals. 
A recent investigation by OpenAI improved CoT as part of a reinforcement learning loop~\cite{o1_learning_to_reason}. 
Kazemi et al. tested LLMs' capability of using `top down' or `backward chain' reasoning \cite{Kazemi2023LAMBADA}. 
They demonstrate a significant performance improvement over CoT and Selection Inference techniques \cite{Kazemi2023LAMBADA}. 
In our work we explicitly teach LLMs to use bottom up, top down and magic set transformation \cite{Bancilhon1985Magic} reasoning techniques, see Figure \ref{fig:reasoning}.

\vspace{-0.55cm}
\vspace{-0.75cm}

\subsection{ATP Reasoning Strategies}

Automated Theorem Proving is one of the oldest subfields of AI, originating in the 1950s. Automated Theorem Provers (ATPs) use logical reasoning and search algorithms to explore
the space of possible proofs and identify those that are valid. Unfortunately, early systems could only be applied to toy problems, mainly for lack of techniques to deal with combinatorially exploding search spaces. The \emph{resolution calculus} developed since the 1960s~\cite{robinson_machine-oriented_1965} marked a first breakthrough in this direction.  
Contemporary ATPs implement strategies and heuristics that make them applicable to realistically-sized 
problems while not compromising on soundness or completeness of solutions (proofs). 
A natural research question is if LLMs can take advantage of ATP search strategies to improve their reasoning using in-context learning. 

Two concepts that are routinely used in ATPs and related systems are bottom up and top down reasoning, also referred to as forward chaining and backward chaining respectively. See Figure \ref{fig:reasoning} and \cite{Harrison2009Handbook} for an overview. 
Bottom up begins a logical deduction process from basic facts and rules and iteratively derives conclusions until it has arrived at the answer to a query. 
It is realised in the hyper-resolution calculus~\cite{overbeek_implementation_1975} and hyper tableaux~\cite{baumgartner_hyper_1996}. 
Moreover, computing materialised views in relational database management systems is a bottom up process.
Bottom up is the primary form of reasoning used by LLMs when performing CoT reasoning \cite{Kazemi2023LAMBADA}.

By contrast, top down reasoning systems start with a query and use rules to recursively derive sub-goals until the sub-goals and query can be proved or disproved by the provided facts. 
Top down reasoning starts with the query to guide the search for a proof. 
Prominent ATP examples are the model elimination calculus \cite{loveland_mechanical_1968} and the Prolog programming language \cite{colmerauer_birth_1996}.


In addition to top down and bottom up reasoning we also explore the use of a magic set transformation approach with LLMs \cite{Bancilhon1985Magic}. Magic set transformation approaches first use a top down exploration to determine the set of rules and facts which are relevant to a query. This often allows the subsequent bottom up reasoning to explore a smaller search space.

While we use the above strategies for a shallow embedding of reasoning, we also investigate coupling ATPs with LLMs in a deep embedding approach.  

A specific type of ATP is a Boolean Satisfiability (SAT) solver. The task of SAT solvers can be defined as checking if the argument from a set of formulas $\{p_1,p_2...,p_n \}$ to another formula $q$ is valid \cite{Biere2021Handbook}. Through the use of negation $\neg$, such an argument is only valid if the set $\{p_1,p_2...,p_n,\neg q \}$ is inconsistent. Satisfiability Modulo Theory (SMT) solvers generalise this problem to reason with First Order Logic (FOL), functions, and arithmetic operations. 

In this paper we do not explore the cutting edge of SAT/SMT solvers, but use an SMT solver, Z3 \cite{Z3}, to perform satisfiability checks for FOL programs of up to approximately twenty rules and facts. We direct the reader to `The Handbook of Satisfiability' \cite{Biere2021Handbook} for a thorough introduction to SAT, and to `Satisfiability Modulo Theories: An Appetizer' \cite{deMoura2009SMT} for an introduction to SMT. 

\section{Methodology}
\label{sec:methods}


In this paper we investigate a total of twelve large language models. These LLMs fall into two categories. The first is frontier LLMs. At any point in time, the frontier LLMs are the models which perform among the highest on standard benchmarks. In this study we look at three frontier LLMs from each of three time periods separated by approximately 9 months:
\begin{itemize}
    \item December 2023 - GPT3 \cite{Brown2020Language}, GPT4 \cite{Openai2023GPT4} and Gemini \cite{Google2023Gemini}.
    \item September 2024 - GPT4o \cite{GPT4o}, Claude Sonnet 3.5 \cite{Anthropic2024Claude}, Llama 3.1 405B \cite{Meta2024Llama}. 
    \item June 2025 - DeepSeek R1 \cite{DeepSeekR1}, GPT-o3 \cite{GPTo3}, Gemini 2.5Pro \cite{Gemini25}
\end{itemize} 

In addition to investigating frontier Large Language models, we investigate small open-source models which are currently available on Ollama. We define small as 15 billion parameters or less. We investigate three small LLMs from June 2025; Phi4 \cite{Phi4}, Gemma3 12B \cite{Gemma3} and Falcon 10B \cite{Falcon3}.

The LLMs used in this study were accessed via different methods. GPT3, GPT-4, GPT-4o and GPT-o3 were accessed through Microsoft Azure endpoints. 
Gemini and Gemini 2.5 Pro were accessed directly using an official API. 
Llama3.1 405B, DeepSeek and Claude 3 Opus model were accessed via an AWS Bedrock API. 
Because different hosting platforms were used, latency was not measured or considered in this study. 

There were some instances where errors in APIs caused a model to return a blank response. 
There were also instances where model token limits were reached resulting in half-complete responses. 
Where these were caught in sufficient time, the experiment was re-run. 
There were also institutional resource limits which restricted the experiments on the 2025 recent frontier models. 
This combination of factors resulted in less than 300 experimental trials for some methods. 

\subsection{Experimental Design}
\label{sec:Experimental_Design}
Our study focuses on Steamroller problems; toy examples used by philosophers to test and extend their reasoning capabilities. 
We use the PrOntoQA benchmark \cite{Saparov2023Language} to generate $300$ problems with $100$ examples requiring one, two and three steps of reasoning. 
In order to challenge the models we explore the so-called `False Ontology' (where statements do not reflect reality) with distractors. 
These are the hardest conditions for reasoning and were not tested in their original paper \cite{Saparov2023Language}. 

As these are true or false questions, an LLM could gain a score of approximately 50\% by randomly guessing. In order to determine the range in which results become statistically significantly different to random guessing we use the Wilson Score Interval \cite{Wilson1927}, see Equation \ref{eqn:confidence}.
\begin{equation}
    p_{\pm} = \frac{np + \frac{1}{2} Z^2}{n + Z^2} \pm \frac{Z}{n + Z^2} \sqrt{p(1-p)n+\frac{Z^2}{4}}
    \label{eqn:confidence}
\end{equation}
where $p_{\pm}$ represent the top and bottom of the interval, $p$ is the portion of the model answers which are correct, $n$ is the number of problems attempted, and $Z$ is the normal interval half width. We use $Z=3$, so that ranges represent confidence at the $99.7\%$ level. For random guessing we assume $p=0.5$ and $n=300$.

We used six different prompts which we named `Normal' (baseline), (Zero-shot) `CoT',  `One-shot CoT', `Bottom Up', `Top Down', and `Magic Set Transformation'. 
The exact prompts used for each of these conditions can be found in Appendix \ref{sec:prompts}. 
To determine accuracy we compare the final instance of `True' or `False' in the model's response to the ground truth. 
We assume default negation; if the model does not give a clear answer, or states that the answer cannot be determined, we assume that it answered false, giving the model a 50\% chance of being correct if it refused to give an answer. 
In some cases the model expressed its answers in strange ways, where it would express the correct answer and then go on to make some remark with the word true or false. 
In order to reduce false negatives, incorrect answers were manually checked. 

Rather than classifying LLM errors into categories \cite{McGinness2024Automated,Xu2023Large}, we compare the LLMs' responses to the  PrOntoQA `Golden Chain of Thought'. Golden CoT is an ordered list of all rules and facts required to reach a conclusion using a bottom up strategy. 
As shown in Figure \ref{fig:reasoning}, top down requires the Golden CoT to be reversed, while magic set requires a reverse pass followed by a forward pass. 
If a model's reasoning contained all required reasoning steps, then its reasoning is said to be complete. 
We use Algorithm \ref{alg:faithfulness} to determine the faithfulness of a model's response;  all of the reasoning steps needed to appear in the correct order. 

\vspace{-0.25cm}
\begin{algorithm}[H]
\caption{Check Faithfulness}
\label{alg:faithfulness}
\begin{pseudo}*
  \hd{Check\_Faithfulness}(L_G,L_M) \\*&
  \textbf{Input:} \textnormal{$L_G$ an ordered list of required facts and rules (Golden Chain of Thought),}\\ *&
  \textnormal{$L_M$ and ordered list of detected facts and rules in LLM response} \\*&
  \textbf{Output:} \textnormal{Faithfulness: Boolean}\\*&
  $L[1]$ denotes the first entry of an ordered list, L.\\*&
  Del$_1(L)$ denotes removing the first entry from an ordered list, L. \\[1ex]
  \textbf{while} $L_M \neq \emptyset$ \textbf{do} \\+
    \textbf{if} $L_M[1] = L_G[1]$ \textbf{do}\\+
        Del$_1(L_G)$ \\
        \textbf{if} $L_G = \emptyset$ \textbf{do}\\+
            \textbf{return} True \\-
        Del$_1(L_M)$ \\
        \textbf{if} $L_M = \emptyset$ \textbf{do}\\+
            \textbf{return} False \\--
    \textbf{else do} \\+
    Del$_1(L_M)$ \\
        \textbf{if} $L_M = \emptyset$ \textbf{do}\\+
            \textbf{return} False \\--
  \end{pseudo}
\end{algorithm}
\vspace{-0.5cm}

Unfortunately there is no automated fully reliable way to detect occurrences of the required rules and facts in the natural language. Previous studies \cite{Saparov2023Language} used a recursive-descent parser using the simple grammar to verify model reasoning. Unparseable proofs were marked as incorrect. To avoid this issue we used a tedious, semi-automated approach where regular expressions were developed to automatically detect patterns of correctly expressed reasoning. As responses were manually reviewed, new patterns were added, reducing future work. A function was also introduced to identify the labeling of statements. 
In the end, 835 responses were reviewed manually and 118 regular expressions were developed to catch different ways of phrasing the facts, rules and queries. The regular expressions were intentionally specific to reduce the false positive rate. 
Once a response was reviewed, its completeness and faithfulness were set by the manual reviewer in order to correctly classify examples where a model's expression of facts and rules were too difficult to identify with regular expressions.

\subsection{Z3 Approach}
We also perform an experiment where we investigate the deep reasoning capability of small LLMs. This requires LLMs to translate the problem into a standardised format for first-order logic. 
This greatly simplifies the task in a way that takes advantage of the strength of LLMs to recognise patterns rather than relying on their reasoning capabilities. 

The small LLMs were given prompts (see Appendix \ref{sec:prompts}) to translate the PrOntoQA problems into
the the standard format 
which includes facts, rules and one query. 
A parser was then used to convert these into a Z3 program in the 
SMT-LIB standard~\cite{barrett_smt-lib_2025} syntax.

There were three experimental conditions:
\begin{enumerate}
\item No Repair - Three examples were given to the LLM as in-context learning. The resulting program was given to Z3 to solve and the Sat/Unsat response used to determine the truth of the query.
\item With Repair - Three examples were given to the LLM. The program is given to Z3 twice, once with a negated query. This results in two Sat/Unsat results. If these were the same there must be an error in the LLMs translation. The LLM would then be given the repair prompt and asked to fix its response. 
\item One Example - Identical to `With Repair' except the LLM is given only one example instead of three. 
\end{enumerate}

The small LLMs used in combination with Z3 in the experiments were run on a local computer with an NVIDIA GeForce RTX 4060Ti GPU, with 16GB of memory. This was able to run the models of up to 14B parameters entirely from the GPU memory.  

We approximate the computational cost of running the open source LLMs in this paper by calculating the number of FLOPs that it takes to run these models. A full derivation of the formulas used can be found in Appendix \ref{sec:app_computational_cost}. We compare results of the `exact' formula to a common approximation for the total number of FLOPs $2 N n$ where $N$ is the number of (active) parameters in the model and $n$ is the total number of prompt and completion tokens \cite{Hoffman2022Training,Kaplan2020Scaling}.

\section{Results}
\label{sec:results}

\begin{figure}[h!]
    \centering
    \includegraphics[width=0.8\textwidth]{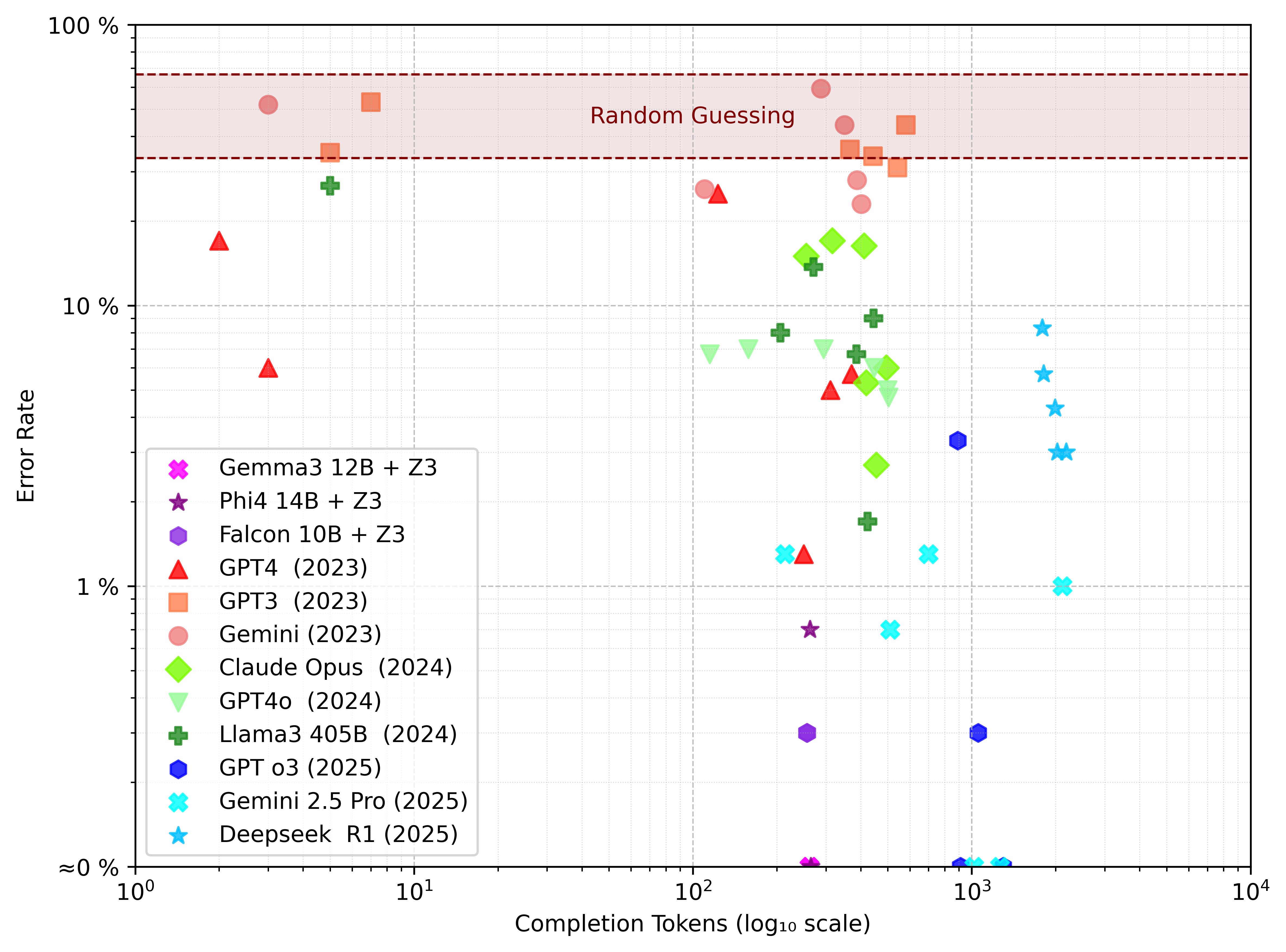}
    \caption{Log-Log scatterplot of error rate against number of completion tokens for each model and experimental condition. The red horizontal lines show the $Z=3$ Wilson Score confidence interval for random guessing.}
    \label{fig:model_error_plot}
\end{figure}

Raw results for frontier model accuracy (correct percentage), reasoning completeness and faithfuless can be found in Table \ref{tab:raw_LLM_data} in Appendix \ref{sec:raw_results}. These results are also displayed in Figure \ref{fig:model_error_plot}. For the normal condition, the December 2023 models usually just wrote `True' or `False' as their answer. As a result they have the lowest completion token count and accuracy. The completion token counts for the 2024 models are approximately two orders of magnitude higher for the normal condition indicating that they have either been trained or internally prompted to automatically perform CoT reasoning. 
The 2025 thinking models are even more verbose, using more than a thousand tokens in most experimental conditions. 

The accuracy results for the December 2023 models are consistent with previous studies; the performance of a model is enhanced by specific instructions, examples and reasoning techniques. By contrast GPT4o, Deepseek R1, Claude Opus, Gemini 2.5 Pro, and GPT-o3 show significantly less change in their number of completion tokens and accuracy when instructed to reason step by step. This is consistent with the hypothesis that these models now have CoT reasoning built into their behaviour during training and in hidden prompts. 

In December 2023, GPT4 was clearly the highest performing model and its scores were significantly higher than GPT3 and Gemini-Pro. In August 2024 and June 2025, the frontier models were much more similar in their capabilities and their scores are reasonably close 
for the last three experimental conditions. 
The small models assisted by Z3, Gemini 2.5 Pro, and GPT-o3 were able to achieve near perfect performance. 
This raises the concern that contamination may be affecting the results. 
The results in Appendix \ref{sec:contamination} reveal that this is not the case, except for Falcon3 10B. 

For open source models, the average number of inference FLOPs for each test condition were calculated. See Figure \ref{fig:FLOPs_plot} and Table \ref{tab:FLOP_data} in Appendix \ref{sec:raw_results}. The small models assisted by Z3 were able to produce higher accuracy results with at approximately 20\% of the computational cost of Deepseek. 
Despite Deepseek producing a very large number of tokens in its responses, it used less compute than Llama405B. 
This is because Deepseek is a mixture of experts model with a smaller number of active parameters. 

\begin{figure}[h!]
    \centering
    \includegraphics[width=0.8\textwidth]{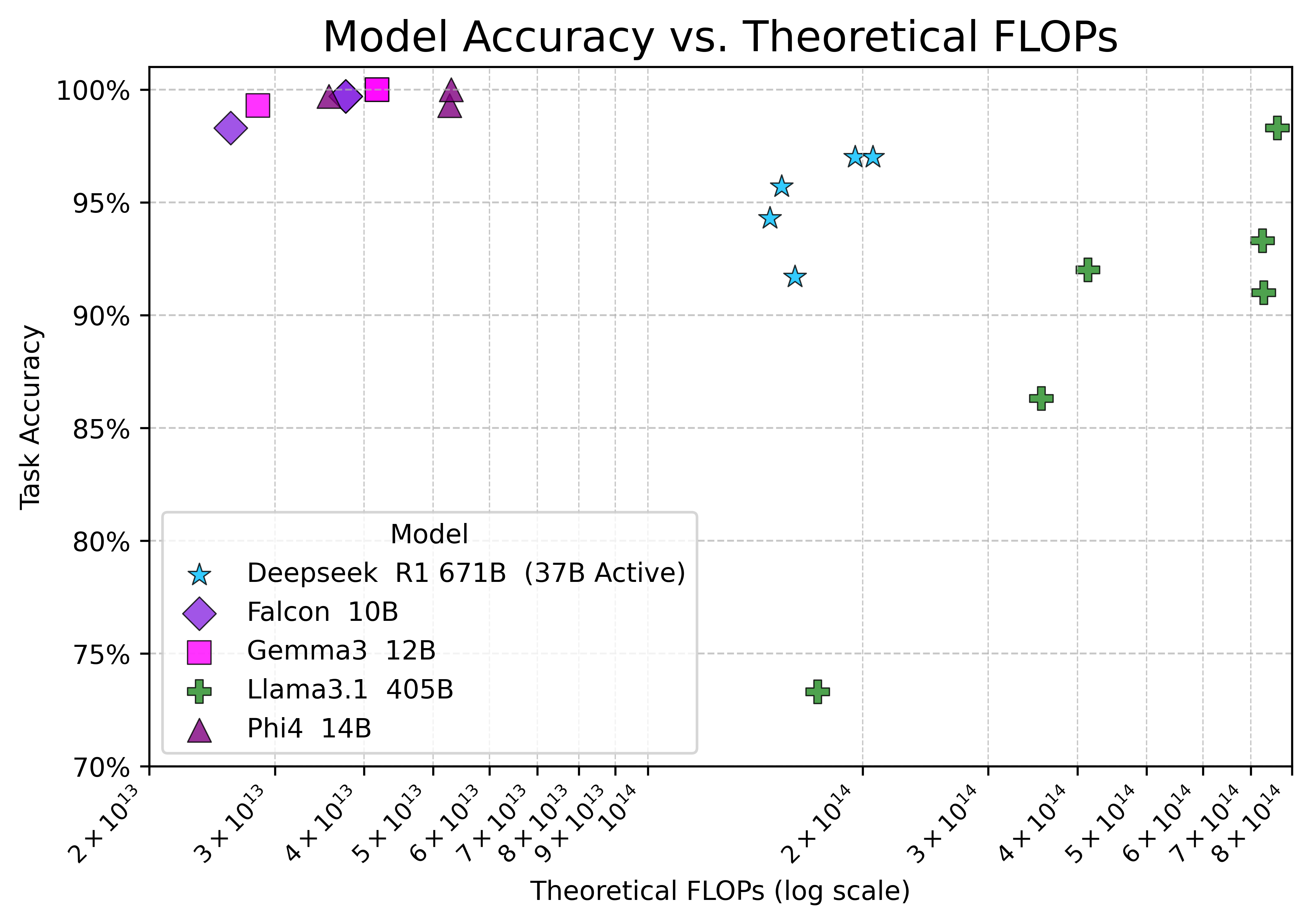}
    \caption{Scatterplot of accuracy against number of FLOPs spent.}
    \label{fig:FLOPs_plot}
\end{figure}

The CPU time spent running Z3 programs  was measured to determine its contribution to the overall cost. 
On average, $300$-$400$ milliseconds of CPU time was used for both the query and its negation with most of this time going to startup, problem parsing and building. Only $1$-$10$ milliseconds of CPU time was used by Z3 to solve each example. This computational cost is negligible compared to the GPU time required to run the small LLMs.

\section{Discussion}
\label{sec:discussion}

The results of these experiments contain a number of significant results. A comparison of the model accuracies in Table \ref{tab:raw_LLM_data} shows that the only significant improvement in 2024 frontier models compared to GPT4 occurs in the zero shot cases. 
The token counts for the normal condition indicate that GPT4o and Claude Opus automatically engage in CoT reasoning. 
This is likely through hidden prompts, as the open source model from the same time period (Llama3.1 405B) did not have hidden prompts and produced only a few tokens for the normal condition. 
This indicates that most of the improvement in LLM performance between December 2023 and September 2024 could be attributed to prompt engineering.

Between September 2024 and June 2025, there was a paradigm shift to thinking models, which are explicitly trained to `think' before responding. 
It appears that this innovative technique provided the capability for LLMs to imitate reasoning as both Gemini 2.5 Pro and GPT-o3 obtained near perfect scores on the ProntoQA dataset.

As the models used in this study were called through a range of APIs there are essentially no measures that could be used to track the computational expense of running the closed source models. If the model architectures (and hidden prompts) were published then it would be possible to calculate the approximate number of FLOPs used for inference. Instead we calculate the computational cost for only the open source models. The $2Nn$ formula was shown to be within $10\%$ of the `exact' theoretical FLOP for all experiments. 

The potential energy consumption and therefore cost, carbon emissions and environmental impact of LLMs are a significant concern \cite{Bender2021StochasticParrots,Luccioni2024Estimating,Luccioni2024Power}. The small LLM results are an important proof of concept that show that combining LLMs with symbolic AIs, including automated reasoners, and other external tools can lead to competitive performance at a significantly reduced computational cost. 
LLMs are good at recognising nuance in language and converting problems to a machine interpretable format. An important area for future work is to identify problems that could leverage this ability to help LLMs interface with other tools.

\section{Conclusions and Future Work}
\label{sec:conclusions}
This study compares the logical deductive reasoning ability of frontier Large Language Models (LLMs) from December 2023, September 2024 and June 2025. We use in-context learning to teach LLMs to apply Automated Theorem Prover (ATP) reasoning strategies to solve the most challenging problems from the PrOntoQA Benchmark. We measured the accuracy of the LLM responses and evaluated the faithfulness of LLM responses to ATP strategies. 

Our results show that the recent training of models to automatically engage in Chain of Thought (CoT) and hidden system prompt engineering strategies improved in zero-shot performance of frontier LLMs from 2023 to 2024. 
However, the improvement in reasoning capabilities seemed to have stalled with no 2024 model performing significantly better than GPT4 (from December 2023) for any other experimental condition. 
The introduction of thinking models in 2025 enabled near perfect accuracy and ability to imitate reasoning strategies in frontier LLMs. 
Similar performance was also enabled at a fraction of the computational cost by combining significantly smaller language models with Z3, an SMT solver.

The steamroller problems explored in this paper are toy examples that could very easily be solved by an SMT solver. 
Another key area for future work would be to apply LLMs to much more difficult ATP benchmarks and test LLM performance using other more sophisticated ATP reasoning strategies. 
Applying the neuro-symbolic approach that combines LLMs with SMT solvers to real world problems with practical value is an important next step to demonstrate the value of this approach. 


{
\raggedbottom
\bibliographystyle{splncs04}
\bibliography{SteamRollerPaper}
}
\let\cleardoublepage\clearpage

\appendix

\section{Appendix - Raw Results}
\label{sec:raw_results}
\begin{table}[H]
\centering
\resizebox{0.75\textwidth}{!}{%
\begin{tabular}{cccccccc}
\hline
 &  &  &  & \textbf{Complete} &  & \textbf{Average} & \textbf{Average} \\
\textbf{Model} & \textbf{Method} & \textbf{Completed} & \textbf{Correct} & \textbf{ Reasoning} & \textbf{Faithfulness} & \textbf{Prompt} & \textbf{Completion} \\
\textbf{} & \textbf{} & \textbf{Examples} & \textbf{Percentage} & \textbf{Percentage} & \textbf{Percentage} & \textbf{Tokens} & \textbf{Tokens} \\
\hline
\hline
\multirow{4}{*}{\centering\shortstack{GPT o3\\ (2025)}}
 & Normal & 283 & 96.7\% & 99.3\% & NA & 114 & 891 \\
 & CoT & 280 & 99.7\% & 100\% & NA & 159 & 1055 \\
 & One Shot CoT & 289 & 100\% & 100\% & NA & 291 & 912 \\
 & Bottom Up & 123 & 100\% & 100\% & 100\% & 493 & 1295 \\
\hline
\multirow{6}{*}{\centering\shortstack{Gemini 2.5\\ Pro (2025)}}
 & Normal & 300 & 98.7\% & 100\% & NA & 105 & 214 \\
 & Chain of Thought & 300 & 98.7\% & 100\% & NA & 150 & 700 \\
 & One Shot CoT & 300  & 99.3\% & 100\% & NA & 284 & 510 \\
 & Bottom Up & 278 & 100\% & 100\% & 99.26\% & 533 & 1262 \\
 & Top Down & 271 & 100\% & 100\% & 99.6\% & 489 & 1019 \\
 & Magic Set & 298 & 99\% & 100\% & 100\% & 439 & 2113 \\
\hline
\multirow{6}{*}{\centering\shortstack{Deepseek \\ R1 (2025)}}
& Normal & 300 & 94.3\% & 99\% & NA & 116 & 1812 \\
 & Chain of Thought & 300 & 95.7\% & 99.7\% & NA & 161 & 1990 \\
 & One Shot CoT & 300  & 91.7\% & 100\% & NA & 299 & 1790 \\
 & Bottom Up & 300 & 97\% & 100\% & 100\% & 506 & 2025 \\
 & Top Down & 300  & 97\% & 99.7\% & 96.7\% & 494 & 2180 \\
\hline
\multirow{6}{*}{\centering\shortstack{Claude Opus \\ (2024)}}
& Normal & 300 & 83\% & 77\% & NA & 131 & 316 \\
 & Chain of Thought & 300 & 83.7\% & 83.33\% & NA & 177 & 411 \\
 & One Shot CoT & 300 & 85\% & 91\% & NA & 318 & 255 \\
 & Bottom Up & 300 & 97.3\% & 97\% & 96.33\% & 535 & 455 \\
 & Top Down & 300 & 94.7\% & 96\% & 79.33\% & 529 & 418 \\
 & Magic Set & 300 & 94\% & 94\% & 93\% & 497 & 494 \\
\hline
\multirow{6}{*}{\centering\shortstack{GPT4o \\ (2024)}}
& Normal & 299 & 93.3\% & 90.6\% & NA & 115 & 300 \\
 & Chain of Thought & 299 & 93\% & 93\% & NA & 158 & 430 \\
 & One Shot CoT & 300 & 93\% & 99.3\% & NA & 294 &  397 \\
 & Bottom Up & 300 & 95.3\% & 99.67\% & 99.7\% & 503 & 706 \\
 & Top Down & 300 & 95\% & 97.33\% & 72\% & 499 & 436.5 \\
 & Magic Set & 299 & 94\% & 94\% & 94\% & 447 & 597 \\
\hline
\multirow{6}{*}{\centering\shortstack{Llama3.1 405B \\ (2024)}}
& Normal & 300 & 73.3\% & 59.7\% & NA & 128 & 87.5 \\
 & Chain of Thought & 300 & 86.3\% & 80.3\% & NA & 171 & 271 \\
 & One Shot CoT & 300  & 92\% & 94.7\% & NA & 308 & 206 \\
 & Bottom Up & 300 & 98.3\% & 97.3\% & 97.3\% & 522 & 423 \\
 & Top Down & 300 & 93.3\% & 90\% & 68\% & 514 & 386 \\
 & Magic Set & 300 & 91\% & 91.3\% & 87\% & 459 & 445 \\
\hline
\multirow{6}{*}{\centering\shortstack{GPT4 \\ (2023)}}
& Normal & 299 & 83\% & 1\% & NA & 110 & 2 \\
 & Chain of Thought & 300 & 75\% & 91.7\% & NA & 162 & 123 \\
 & One Shot CoT & 300  & 94\% & 95\% & NA & 301 & 92.3 \\
 & Bottom Up & 300 & 95\% & 96.3\% & 94\% & 510 & 311 \\
 & Top Down & 300  & 98.7\% & 96\% & 45.7\% & 499 & 250 \\
 & Magic Set & 300 & 94.3\% & 98.7\% & 96.3\% & 444 & 371 \\
\hline
\multirow{6}{*}{\centering\shortstack{GPT3 \\ (2023)}}
& Normal & 300 & 47\% & 9\% & NA & 110 & 19.7 \\
 & Chain of Thought & 300 & 65\% & 67.7\% & NA & 95 & 461.5 \\
 & One Shot CoT & 300 & 66\% & 77.3\% & NA & 301 & 443 \\
 & Bottom Up & 291 & 69\% & 60.5\% & 56\% & 510 & 542 \\
 & Top Down & 296 & 56\% & 65.9\% & 45\% & 499 & 581 \\
 & Magic Set & 299 & 64\% & 72.2\% & 70.6\% & 444 & 366 \\
\hline
\multirow{6}{*}{\centering\shortstack{Gemini Original \\ Release (2023)}}
& Normal & 294 & 48\% & 0\% & NA & 94 & 3 \\
 & Chain of Thought & 290 & 40.7\% & 88\% & NA & 148 & 288 \\
 & One Shot CoT & 300 & 74\% & 60\% & NA & 283 & 110 \\
 & Bottom Up & 296 & 72\% & 66.9\% & 58.8\% & 497 & 388 \\
 & Top Down & 300 & 56\% & 72.33\% & 48\% & 485 & 350 \\
 & Magic Set & 300 & 77\% & 70.7\% & 68\% & 431 & 402 \\
\hline
\end{tabular}%
}
\caption{Number of examples, accuracy, and fraction of cases with complete and faithful reasoning for all frontier model experiments. The complete reasoning and faithfulness percentages are reported as a fraction of the examples that were correct. The average number of prompt and completion tokens for each experiment is also included.}
\label{tab:raw_LLM_data}
\end{table}

\clearpage

\begin{table}[htbp]
\centering
\resizebox{\textwidth}{!}{%
\begin{tabular}{clccccc}
\hline
\textbf{Model} & \textbf{Method} & \textbf{\shortstack{Average \\ Total Tokens}} & \textbf{\shortstack{Theoretical \\ FLOPs}} & \textbf{\shortstack{Approximation \\ FLOPs}} & \textbf{\shortstack{Percentage FLOP \\ Discrepancy}} & \textbf{\shortstack{Task \\ Accuracy}} \\
\hline
\hline
\multirow{5}{*}{\centering\shortstack{Deepseek \\ R1 671B \\ (37B Active)}}
& Normal & 1928 & $1.483\times10^{14}$ & $1.419\times10^{14}$ & 4.3\% & 94.3\% \\
 & Chain of Thought & 2151 & $1.54\times10^{14}$ & $1.658\times10^{14}$ & 4.6\% & 95.7\% \\
 & One Shot CoT & 2090 & $1.607\times10^{14}$ & $1.538\times10^{14}$ & 4.3\% & 91.7\% \\
 & Bottom Up & 2531 & $1.953\times10^{14}$ & $1.862\times10^{14}$ & 4.7\% & 97\% \\
 & Top Down & 2674 & $2.068\times10^{14}$ & $1.968\times10^{14}$ & 4.9\% & 97\% \\
\hline
\multirow{6}{*}{\centering\shortstack{Llama3.1 \\ 405B }}
& Normal & 216 & $1.73\times10^{14}$ & $1.75\times10^{14}$ & 1.0\% & 73.3\% \\
& Chain of Thought & 442 & $3.56\times10^{14}$ & $3.58\times10^{14}$ & 0.5\% & 86.3\% \\
 & One Shot CoT & 514 & $4.14\times10^{14}$ & $4.16\times10^{14}$ & 0.6\% & 92\% \\
 & Bottom Up & 945 & $7.63\times10^{14}$ & $7.65\times10^{14}$ & 0.3\% & 98.3\% \\
 & Top Down & 900 & $7.27\times10^{14}$ & $7.29\times10^{14}$ & 0.3\% & 93.3\% \\
 & Magic Set & 904 & $7.3\times10^{14}$ & $7.32\times10^{14}$ & 0.3\% & 91\% \\
\hline
\multirow{3}{*}{\centering\shortstack{Phi4 \\ 14B }}
& 3 Examples  & 1888 & $5.27\times10^{13}$ & $5.29\times10^{13}$ & 0.4\% & 99.3\% \\
& 3 Examples 
+ Repair & 1899 & $5.3\times10^{13}$ & $5.32\times10^{13}$ & 0.4\% & 100\% \\
& 1 Example 
+ Repair & 1281 & $3.57\times10^{13}$ & $3.59\times10^{13}$ & 0.6\% & 99.7\% \\
\hline
\multirow{3}{*}{\centering\shortstack{Gemma3 \\ 12B }}
& 3 Examples  & 1900 & $4.17\times10^{13}$ & $4.56\times10^{13}$ & 9.4\% & 100\% \\
& 3 Examples 
+ Repair & 1900 & $4.17\times10^{13}$ & $4.56\times10^{13}$ & 9.4\% & 100\% \\
& 1 Example 
+ Repair & 1293 & $2.84\times10^{13}$ & $3.1\times10^{13}$ & 9.2\% & 99.3\% \\
\hline
\multirow{3}{*}{\centering\shortstack{Falcon \\ 10B }}
& 3 Examples  & 1938 & $3.77\times10^{13}$ & $3.88\times10^{13}$ & 2.9\% & 99.7\% \\
& 3 Examples 
+ Repair & 1941 & $3.77\times10^{13}$ & $3.88\times10^{13}$ & 2.9\% & 99.7\% \\
& 1 Example 
+ Repair & 1342 & $2.6\times10^{13}$ & $2.68\times10^{13}$ & 3.2\% & 98.3\% \\
\hline
\end{tabular}%
}
\caption{Raw Data by model and method.}
\label{tab:FLOP_data}
\end{table}

\section{Appendix Derivations of Computational Cost}
\label{sec:app_computational_cost}

The task to be completed by a generative transformer is as follows: Let $V$ be a vocabulary consisting of $n_{\text{vocab}}$ unique tokens, $V=\{x_1,x_2,...,x_n\}$ with one of the tokens $x_s$ being the stop token. 
Let $\mathbf{y}= (y_1, y_2, ... ,y_L)$ be an arbitrary sequence of tokens from $V$ and the number of elements in $\mathbf{y}$ be $L=|\mathbf{y}|$. 
Let $p_\theta$ be a probability distribution over all tokens in $V$ parametrised by $\theta$. 
The task of the transformer is to model the joint probability of a token sequence:
$p_\theta(\mathbf{y}) = \Pi_{i=1}^L p_\theta(y_{i}|y_1,y_2,...,y_{i-1})$

\subsection{Transformer Architecture}
The transformer architecture was originally published in 2017 \cite{AttentionisAllYouNeed} which contained both encoder and decoder stacks. However most Large Language Models are pre-layer-normalisation, decoder-only Transformers \cite{Brown2020Language,Openai2023GPT4}. 
The decoder-only transformer consists of several components. The first is an embedding matrix\footnote{The embedding matrix must have dimension $n_{\text{vocab}}\times d_{\text{model}}$} which converts input tokens to vector embeddings. 
Then a positional encoder adds either fixed sinusoidal or learned values, to each vector embedding to indicate its position in the sequence. 

The multi-head attention block then begins with layer normalisation before a multi-head attention. 
The block contains matrices $W_{Q_h}$, $W_{K_h}$ and $W_{V_h}$ where $h$ indicates corresponding to the hth attention head. 
The (positionally encoded) token embeddings from the previous layer matrices are multiplied by these matrices to create the query, key and value matrices, $Q_h$, $K_h$ and $V_h$ 
\footnote{Each of the $W_{Q_h}$, $W_{K_h}$ and $W_{V_h}$ matrices must have dimension $d_{\text{model}} \times d_{\text{attn}}$. As there are $n_{\text{heads}}$ of each this can be concatenated into one $W_Q$, $W_K$ and $W_V$ for more efficient computation. Usually $d_{\text{model}}=n_{\text{heads}}d_{\text{attn}}$ in which case $W_Q$, $W_K$ and $W_V$ have size $d_{\text{model}} \times d_{\text{model}}$.}. 
$Q_hK_h^\top$ is calculated to create a matrix which contains scores indicating how much each token should attend to every other token \cite{Liu2018Generating}. 
In the decoder only architecture, a mask is applied to the top-right triangle of this matrix, converting the values to $-\infty$ to prevent tokens to attending to future tokens \cite{Liu2018Generating,AttentionisAllYouNeed}. 
Softmax is then applied before multiplying by the value matrix $V_h$ to create $h$ output matrices $O_h$ \cite{AttentionisAllYouNeed}. 
These output matrices are concatenated and passed through a linear layer to create the final output from the attention layer. 

At this stage most decoder-only transformers have a post residual connection layer which adds the embedding that was passed into the attention block to the output of the attention layer before the next block's pre-layer normalisation \cite{Xiong2020Layer}. 
The same process is used to add the input of a feed forward network block to its output \cite{Xiong2020Layer}. 
The feed forward network increase the dimension of the embedding, apply non-linear activation functions to each entry, before contracting back to the model dimension \cite{AttentionisAllYouNeed}. 
Then there is a final residual connection layer.

These multi-head attention and feed forward blocks combine to form a layer. 
Typical LLMs have between 6 and 200 layers depending on their size. 
After all of the layers there is a final normalisation layer and then a linear layer uses a weight matrix to project the output into the vocabulary space \cite{AttentionisAllYouNeed}. A common parameter efficient implementation called tied input-output embeddings uses the transpose of the embedding matrix to obtain the final embedding \cite{Press2017Using}. 
This gives each token a score for each entry in the vocabulary, which a final softmax function converts to a probability distribution for the next token \cite{AttentionisAllYouNeed}.


\subsection{Theoretical FLOPs Calculations}
In this paper we calculate the theoretical FLOPs required for the open source models used, DeepSeek R1, Llama-3 405B, Phi-4, Gemma-3 12B and Falcon-3 10B. Note that these models with the exception of Phi-4 do not use biases in their linear layers. 
As this only contributes a small number of parameters to the model and FLOPs at inference we will ignore these for all models. 
These models all use RoPE \cite{RoPE} for positional embedding and RMSNorm \cite{RMSNorm} for normalisation. 
DeepSeek uses Multi-Head Latent Attention (MLA) \cite{DeepSeekV2}, however we ignore this technique in our calculations as its primary impact is reducing the memory usage of the KV cache. We acknowledge that for long sequences MLA could reduce the FLOPs and that our calculation does not capture this effect.

Kaplan provides a method for approximating the number of floating point operations (FLOPs) required for a forward pass of a decoder only transformer using the following notation \cite{Kaplan2020Scaling}: 
\define{$d_{\text{model}}$} is the embedding size, also known as the dimension of the residual stream, 
\define{$d_{\text{ff}}$} is the expanded dimension of the feed forward layer, 
\define{$d_{\text{attn}}$} is the dimension of each attention head, given by $d_{\text{model}} = n_{\text{heads}} d_{\text{attn}}$, 
\define{$n_{\text{heads}}$} is number of attention heads in each multi-head attention block, 
\define{$n_{\text{ctx}}$} number of tokens in the input context, 
\define{$n_{\text{max}}$} as the maximum context length, 
\define{$n_{\text{layer}}$} is the number of layers,
\define{$g$} is the Group Query Attention ratio and 
\define{$n_{\text{vocab}}$} is the number of tokens in the vocabulary, $V$, therefore $n_{\text{vocab}} = |V|$. We will use 
\define{$N$} to denote the total number of parameters in the model (model size), which, assuming tied input-output embeddings, RoPE positional embeddings and RMSNorm Normalisation is given by 
$N \approx d_{\text{model}}n_{\text{vocab}}+
n_{\text{layers}}(2d_{\text{model}} +
(1+2/g)n_{\text{heads}}d_{\text{model}}d_{\text{attn}} + 
n_{\text{heads}}d_{\text{model}}d_{\text{attn}} +
3d_{\text{model}}d_{\text{ff}}) + d_{\text{model}}$, 
simplifying gives 
$N \approx d_{\text{model}}n_{\text{vocab}}+
n_{\text{layers}}d_{\text{model}}
(2+ (2+2/g)d_{\text{model}}+3d_{\text{ff}}) + d_{\text{model}}$ \footnote{Note that $n_{\text{max}}d_{\text{attn}}$ values must be precomputed for RoPE but these are not learned parameters and do not count towards the parameter count}.

We will make the same assumption that the computational cost of multiplying two matrices of dimensions $a\times b$ and $b \times c$ is $2abc$ FLOPs \cite{Kaplan2020Scaling}. This is because for each of the $a\times c$ entries in the output matrix we perform $b$ multiplications and $b-1\approx b$ additions. However due to causal masking in attention, the upper diagonal of a square matrix (assuming $a=c$) does not always need to be calculated, in this case the number of FLOPs required for calculation is $ba(a+1)$ FLOPs \cite{FlashAttention}. 

We start by considering the computational cost of the first token output. The total computational cost of processing the entire input sequence to generate the first output token will be given by the $C_1 = C_{1_{\text{embedding}}} + n_{\text{layer}}(C_{1_{\text{attn}}} + C_{1_{\text{ff}}}) + C_{1_{\text{ouput}}}$.

The first step is to convert the context tokens to vector embeddings. As these are one hot encoded this requires only lookup operations and therefore technically do not contribute any FLOPs to the overall computational cost. 
Traditionally position embeddings were added to each of the input token embeddings uses $C_{1_{\text{embedding}}}=n_{\text{ctx}} d_{\text{model}}$ FLOPs (addition operations) \cite{AttentionisAllYouNeed}, before the first attention block. 
This is a very small contribution to the overall computational cost, so some authors ignore this \cite{Kaplan2020Scaling}. 
The models in this study use RoPE which actually applies positional embeddings within the attention blocks \cite{RoPE}. 
Therefore the cost of embedding will be calculated in these sections and there is no cost (in FLOPs) to calculate the initial embeddings, so $C_{1_{\text{embedding}}}=0$ FLOPs.

After the tokens are converted to vector embeddings the first Layer normalisation block is applied. For each embedding vector the RMSNorm calculates the square of each entry ($d_{\text{model}}$ multiplications), sums all the entries ($d_{\text{model}}-1$ additions), divides the sum by the number of entries ($1$ division), takes the square root and inverts the result (approximately $2$ FLOPs), scales each entry by this factor ($d_{\text{model}}$ divisions) and multiplies the result by a learned scaling parameters $g$ ($d_{\text{model}}$ multiplications). This has a total cost of $4d_{\text{model}} +2 \approx 4 d_{\text{model}}$ per input token \cite{Kaplan2020Scaling}. 
Given a context length of $n_{\text{ctx}}$, this results in $4n_{\text{ctx}} d_{\text{model}}$ FLOPs. 

In the attention block first the $Q$, $K$ and $V$ matrices must be computed by multiplying the vector embeddings ($n_{\text{ctx}} \times d_{\text{model}}$) by the $W_Q$, $W_K$ and $W_V$ matrices ($d_{\text{model}} \times d_{\text{model}}$) which requires $6 n_{\text{ctx}}d_{\text{model}}^2$ FLOPs. 
Note that models which use Group Query Attention \cite{Ainslie2023GQA} reduce this cost by reusing (sharing) Key and Value matrices for a $g$ Query matrices. 
In this case the cost is reduced to $(2+4/g) n_{\text{ctx}}d_{\text{model}}^2$ FLOPs.
Then RoPE is applied to the $n_{\text{ctx}}$ rows of the $Q$ and $K$ matrices corresponding to each token. 
A rotation matrix is applied to each pair of elements in the rows in the Q and K matrices. This makes a total of $2n_{\text{heads}} $ vectors to which it is applied. 
The cost of applying the rotation is $3d_{\text{attn}} $FLOPs per entry, two multiplications and one addition per attention dimension. 
Therefore the total cost of RoPE for one head is $6n_{\text{ctx}}d_{\text{attn}}$ FLOPs. Taking into account Query Group Attention and summing over all of the heads gives $(3+3/g)n_{\text{ctx}}d_{\text{model}} $FLOPs.

Computing the elements of $QK^{\top}$ not covered by the causal mask requires $n_{\text{ctx}}(n_{\text{ctx}}+1)d_{\text{model}}$ FLOPs
\footnote{Notice that the true derivation calculates attention for each of $n_{\text{heads}}$ heads of dimension $d_{\text{attn}} =d_{\text{model}}/n_{\text{heads}}$ gives $n_{\text{heads}}(n_{\text{ctx}}(n_{\text{ctx}}+1)d_{\text{attn}})=n_{\text{ctx}}(n_{\text{ctx}}+1)d_{\text{model}}$ which reduces to the same result as the simplified concatenated model.} 
and scaling (element wise division) requires one FLOP per element: $0.5\times n_{\text{heads}}n_{\text{ctx}}(n_{\text{ctx}}+1)$ FLOPs. 
When considering the FLOPs used by softmax function we must consider each of the $n_{\text{heads}}$ attention head separately. 
For each head the softmax is applied to each row of the matrix so all $\frac{n_{\text{ctx}}(n_{\text{ctx}}+1)}{2}$ entries not covered by the mask. 
This leads to a total cost of approximately 10 FLOPs per entry, depending on how Softmax is implemented. 
Therefore the total cost of this across all heads is $5 n_{\text{heads}} n_{\text{ctx}}(n_{\text{ctx}}+1)$.
Multiplying the result (a $n_{\text{ctx}} \times n_{\text{ctx}}$ matrix) by the $V$ ($n_{\text{ctx}} \times d_{\text{model}}$) matrix then uses $n_{\text{ctx}}(n_{\text{ctx}}+1)d_{\text{model}}$ FLOPs as the above diagonal entries of the former matrix are 0. 
Then this output is multiplied by a weight matrix ($d_{\text{model}} \times d_{\text{model}}$) at a cost of $2n_{\text{ctx}}d_{\text{model}}^2$ FLOPs.
Adding the original vector embedding to this output requires $n_{\text{ctx}}\times  d_{\text{model}}$ FLOPs.
Therefore each attention block (including layer normalization) requires $C_{1_{\text{attn}}}=
(4+4/g)n_{\text{ctx}}d_{\text{model}}^2+
(8+3/g)n_{\text{ctx}}d_{\text{model}}+ 
2n_{\text{ctx}}(n_{\text{ctx}}+1)d_{\text{model}} +
5.5 n_{\text{heads}} n_{\text{ctx}}(n_{\text{ctx}}+1)$.

The feed forward layer begins with a Layer normalisation ($4n_{\text{ctx}} d_{\text{model}}$ FLOPs). 
The the models that we use in this paper use either SwiGLU or GeGLU activation functions. 
To take into account both of these we introduce a new parameter \define{$n_{\text{$n_A$}}$} which is the number of FLOPs per application of the activation function.
$n_A \approx 6$ for SwiGLU (SiLU) and $n_A \approx 10$ for GeGLU (GeLU) \cite{SiLU,GeLU}. 
In either case the input matrix ($n_{\text{ctx}} \times d_{\text{model}}$) is multiplied by both a gate matrix $W_{\text{gate}}$ and an up-projection matrix $W_{\text{up}}$. 
Both of these matrices have size ($d_{\text{model}}\times d_{\text{ff}}$) and both matrix multiplications combined cost $4 n_{\text{ctx}} d_{\text{model}} d_{\text{ff}}$ FLOPs. 
Then the SiLU/GeLU activation functions must then be applied to each of the $n_{\text{ctx}} \times d_{\text{ff}}$ entries of the matrix resulting from multiplying the normalised input by $W_{\text{gate}}$, the cost for this is $n_An_{\text{ctx}} d_{\text{ff}}$ FLOPs. 
Then element wise multiplication between the result and the $W_{\text{up}}$ matrix  costs $n_{\text{ctx}} d_{\text{ff}}$ FLOPs. 
Finally multiplication by the $W_{\text{down}}$ matrix requires another $2 n_{\text{ctx}} d_{\text{model}} d_{\text{ff}}$ FLOPs. 
Adding the feed-forward input vector embedding to this output requires $n_{\text{ctx}}\times  d_{\text{model}}$ FLOPs, resulting in a total of $C_{1_{\text{ff}}}=5n_{\text{ctx}} d_{\text{model}} + 
6 n_{\text{ctx}} d_{\text{model}} d_{\text{ff}}+
(n_A+1)n_{\text{ctx}} d_{\text{ff}}$, for the feed forward block. 

To convert the vector embedding into an output over the vocabulary there are three steps. 
First, there is a final RMSNorm layer, however this time only the embedding corresponding to the final token needs to be calculated so the cost is $4 d_{\text{model}}$ FLOPs.
The second step is to multiply this final token embedding ($1 \times d_{\text{model}}$) by the transpose of the embedding matrix ($d_{\text{model}} \times n_{\text{vocab}}$) which costs $2 d_{\text{model}}n_{\text{vocab}}$ FLOPs. 
The third step is to apply the softmax function to the final row\footnote{Note that at inference time the proceeding token probability distribution only needs to be computed for the final input token. 
At training time this softmax would have to be applied to all $n_{\text{ctx}}$ tokens.}, 
which has an approximate cost of $10n_{\text{vocab}}$ FLOPs. Therefore $C_{1_{\text{output}}} = 4d_{\text{model}} + 2 d_{\text{model}}n_{\text{vocab}} + 10n_{\text{vocab}}$.

Therefore the approximate cost for the first token is:
\begin{flalign*}
    C_1 = &C_{1_{\text{embedding}}} + n_{\text{layer}}(C_{1_{\text{attn}}} + C_{1_{\text{ff}}}) + C_{1_{\text{output}}}\\
    C_1 = &0+n_{\text{layer}}[(4+4/g)n_{\text{ctx}}d_{\text{model}}^2+
(8+3/g) n_{\text{ctx}}d_{\text{model}}+ 
2n_{\text{ctx}}(n_{\text{ctx}}+1)d_{\text{model}} +
5.5 n_{\text{heads}} n_{\text{ctx}}(n_{\text{ctx}}+1)] + \\ 
    &n_{\text{layer}} (5n_{\text{ctx}} d_{\text{model}} + 
6 n_{\text{ctx}} d_{\text{model}} d_{\text{ff}}+
(n_A+1)n_{\text{ctx}} d_{\text{ff}}) +\\
    &4d_{\text{model}} + 2d_{\text{model}}n_{\text{vocab}} + 10n_{\text{vocab}}\\
    C_1 = &n_{\text{ctx}}n_{\text{layer}}[d_{\text{model}}([4+4/g]d_{\text{model}}+18+2n_{\text{ctx}}+6d_{\text{ff}}) + 5.5n_{\text{heads}}(n_{\text{ctx}}+1) + (n_A+1) d_{\text{ff}})] + 4d_{\text{model}}\\
    &+ n_{\text{vocab}}( 2d_{\text{model}} + 10)
\end{flalign*}
As long as sufficient memory is available, the cost for each subsequent token can be reduced using KV caching \cite{Dai2019Transformer}. We will say the total computational cost for each token after the first is given by $C_i = C_{i_{\text{embedding}}} + n_{\text{layer}}(C_{i_{\text{attn}}} + C_{i_{\text{ff}}}) + C_{i_{\text{output}}}$. Similar to before $C_{i_{\text{embedding}}}= 0$ FLOPs. 

The layer normalisation cost for this one token is $4d_{\text{model}}$. 
Due to KV caching, only one row needs to be calculated for each Query, Key and Value matrix requiring $(2+4/g)d_{\text{model}}^2$ FLOPs. 
Applying RoPE to the new Q and K rows costs $(3+3/g)d_{\text{model}}$ FLOPs.
Only one new attention row needs to be calculated by taking an inner product of the new Query vector with each of the cached $K$ vectors plus the new $i$th $K$ vector. Therefore the new attention row requires $2n_{\text{heads}}(n_{\text{ctx}}+i)d_{\text{attn}}=2d_{\text{model}}(n_{\text{ctx}}+i)$ FLOPs. 
Scaling and applying the softmax costs an additional $11n_{\text{heads}}(n_{\text{ctx}}+i)$  FLOPs. 
Multiplying by the $V$ matrix costs  $2n_{\text{heads}}(n_{\text{ctx}}+i)d_{\text{attn}}=2d_{\text{model}}(n_{\text{ctx}}+i)$ FLOPs. 
After concatenation the resulting vector will be multiplied by $W_O$ which costs another $2d_{\text{model}}^2$ FLOPs. 
Finally adding the residual has a cost of $d_\text{model}$ FLOPs. Therefore $C_{i_{attn}} = (8+3g)d_{\text{model}}+8d_{\text{model}}^2+ 4(n_{\text{ctx}}+i)d_{\text{model}} + 11n_{\text{heads}}(n_{\text{ctx}}+i)$.

The feed forward layer only needs to act on the ith token, reducing the cost of layer normalisation to $4d_{\text{model}}$, the $W_\text{gate}$ and $W_\text{up}$ multiplications to $2 d_{\text{model}} d_{\text{ff}}$ FLOPs each, the application of activation functions to $n_A d_{\text{ff}}$ FLOPs, element wise multiplication to $d_\text{ff}$, multiplication by the $W_\text{down}$ matrix to $2 d_{\text{model}} d_{\text{ff}}$ and the addition of the residual embedding to $d_{\text{model}}$ FLOPs respectively. Therefore $C_{i_{\text{ff}}}= 5d_{\text{model}} + 6d_{\text{model}} d_{\text{ff}}+(n_A+1)d_{\text{ff}}$. The computational cost of the output layer is unchanged, $C_{i_{\text{output}}}=4d_{\text{model}}+2 d_{\text{model}}n_{\text{vocab}} + 10n_{\text{vocab}}$.

Therefore the approximate cost for the ith token is 
\begin{flalign*}
    C_i = &C_{i_{\text{embedding}}} + n_{\text{layer}}(C_{i_{\text{attn}}} + C_{i_{\text{ff}}}) + C_{i_{\text{output}}}\\
    C_i = &0 + 
    n_{\text{layer}}[(8+3g)d_{\text{model}}+(4+4/g)d_{\text{model}}^2+4d_{\text{model}}(n_{\text{ctx}}+i) + 11n_{\text{heads}}(n_{\text{ctx}}+i)] + \\ 
    &n_{\text{layer}} [5d_{\text{model}} + 6d_{\text{model}} d_{\text{ff}}+(n_A+1)d_{\text{ff}}] +
    4d_{\text{model}}+ 2d_{\text{model}}n_{\text{vocab}} + 10n_{\text{vocab}}\\
    C_i =& n_{\text{layer}}[(13+3/g)d_{\text{model}}+(4+4/g)d_{\text{model}}^2+4d_{\text{model}}n_{\text{ctx}} + 11n_{\text{heads}}n_{\text{ctx}} + 6d_{\text{model}} d_{\text{ff}}+(n_A+1)d_{\text{ff}}] +\\
    &4d_{\text{model}}+ 2d_{\text{model}}n_{\text{vocab}} + 10n_{\text{vocab}}+\\
    & i (4d_{\text{model}} + 11n_{\text{heads}})n_{\text{layer}}
\end{flalign*}

For future calculations we will divide $C_i$ into two components, one that is independent of $i$ called $C_{i0}$, and a second which is proportional to $i$, $iC_{ic}$. 
Therefore $C_i = C_{i0} + iC_{ic}$, $C_{ic}=(4d_{\text{model}} + 11n_{\text{heads}})n_{\text{layer}}$ and $C_{i0}=n_{\text{layer}}[(13+3/g)d_{\text{model}}+(4+4/g)d_{\text{model}}^2+4d_{\text{model}}n_{\text{ctx}} + 11n_{\text{heads}}n_{\text{ctx}} + 6d_{\text{model}} d_{\text{ff}}+(n_A+1)d_{\text{ff}}] +4d_{\text{model}}+ 2d_{\text{model}}n_{\text{vocab}} + 10n_{\text{vocab}}$
If the number of output tokens is length $n_{\text{out}}$ then the total cost of inference is given by 
\begin{flalign*}
    C_\text{total} = &C_1+\sum_{i=2}^{n_\text{out}}C_i\\
    C_\text{total} = &C_1 + (n_\text{out}-1)C_{i0} + \frac{(n_\text{out}+2)(n_\text{out}-1)}{2}C_{ic}
\end{flalign*}

Note that DeepSeek is a mixture of Experts Model. This does not impact the attention layers, but means that there are many sets of expert Feed Forward Network layers, eight of which are used to respond to any given query. 

\section{Appendix - Contamination Check}
\label{sec:contamination}

At the start of the experiment contamination was not a significant risk. The ProntoQA dataset had only just been released and it was not widely used. Now the benchmark is well cited. Also as informal preliminary contamination test, the authors asked frontier models about ProntoQA in 2025, they knew about it and can describe it.

In order to check for contamination we replaced all of the concept names, property families and proper nouns in the default ProntoQA script with ones that do not appear. For example `animal' was replaced with `angel'. Then the highest performing Frontier Model and the small LLMs experiments were run again on the newly generated `original' examples.

\begin{table}[htbp]
\centering
\renewcommand{\arraystretch}{1.3}
\noindent\textbf{Frontier LLM}

\vspace{0.5em}
\begin{tabular}{llcccc}
\toprule
\textbf{Model} & \textbf{Condition} & \multicolumn{4}{c}{\textbf{Accuracy (\%)}} \\
\cmidrule(lr){3-6}
& & Normal & CoT & 1-shot & Bottom \\
& &  &  & CoT & Up \\
\midrule
Gemini Pro 2.5 & Original ProntoQA & 89.7 & 98.7 & 99.3 & 100 \\
Gemini Pro 2.5 & New Variables & 99.3 & 100 & 99.7 & 100 \\
\bottomrule
\end{tabular}
\vspace{2em}

\noindent\textbf{Small LLM + Z3}

\vspace{0.5em}
\begin{tabular}{lcc}
\toprule
\textbf{Model}  & \multicolumn{2}{c}{\textbf{Accuracy (\%)}} \\
\cmidrule(lr){2-3}
 & Original Pronto QA & New Variables \\
\midrule
Phi4 (14B)   & 100 & 99.3 \\
Gemma (12B)  & 100 & 100 \\
Falcon (10B)  & 99.7 & 93.0 \\
\bottomrule
\end{tabular}
\vspace{0.5em}
\caption{Raw data for Gemini 2.5 pro (the highest performing frontier model) and small LLM contamination analysis. The first four methods' prompts were used to test the frontier model. The `Three examples with repair' prompt was used for the small Large Language Models.}
\label{tab:contamination_data}
\end{table}

The results show that there is no significant contamination for Gemini2.5 Pro, Phi4 or Gemma3 12B. However there is a significant reduction in the scores of Falcon3 10B. 

\section{Appendix - Prompts}
\label{sec:prompts}

A concise summary of each of the prompts is included below 

\paragraph{\textbf{Example and Query}}

\begin{quote}
    \footnotesize{\{question\} = \textit{``Each sheep is sunny. Each sheep is a feline. Sheep are mammals. Felines are aggressive. Every feline is a snake. Felines are carnivores. Each snake is luminous. Snakes are cats. Every dog is not luminous. Each snake is an animal. Animals are fast. Carnivores are opaque. Each mammal is floral. Each vertebrate is not feisty. Each vertebrate is a cow. Alex is a sheep. Alex is a vertebrate.''}}
\end{quote}

\begin{quote}
    \footnotesize{\{query\} = \textit{``True or false: Alex is luminous.''}}
\end{quote}

\paragraph{\textbf{Normal prompt}}
\begin{quote}
    \footnotesize{prompt = \textit{``\{question\} \{query\}''}}
\end{quote}

\paragraph{\textbf{Chain of thought prompt}}
\begin{quote}
    \footnotesize{prompt = \textit{``Consider the following statements and the given query. Use your reasoning skills to determine if the query is true or false based on the statements. Explain your thought process step by step as you analyze the relationship between the statements and the query. \\
    Statements: \{question\} \\
    Query: \{query\}''}}
\end{quote}

\paragraph{\textbf{One-shot Chain of thought}}
\begin{quote}
    \footnotesize{prompt = \textit{``Consider the following statements and the given query. Use your reasoning skills to determine if the query is true or false based on the statements. Follow the format of the example question that follows.\\
    Example Statements: `All cats are birds. No bird swims. Whiskers is a cat.' \\
    Query: `True or false: Whiskers swims.' \\
    Example Reasoning: `Let's figure out if Whiskers swims. This is not provided directly in the statements. However it does state that Whiskers is a cat. Then it states that all cats are birds. Therefore Whiskers is a bird. It then states that no bird swims. Since Whiskers is a bird this means that Whiskers does not swim. Therefore the query ``Whiskers Swims'' is false.' \\
    Explain your thought process step by step as you analyze the relationship between the statements and the query. \\
    Statements: \{question\} \\
    Query: \{query\}''}}
\end{quote}

\paragraph{\textbf{Bottom up prompt}}
\begin{quote}
    \footnotesize{prompt = \textit{``Consider the following statements, which include rules and then facts, along with the given query. Facts are have a specific named instance. For example ``Sam is a cat'' is a fact because an explicit name ``Sam'' is given. Rules establish a connection between two general classes without referring to a specific instance. For example ``Cats are birds'' is rule, not a fact because there is no named instance of a specific cat or bird. Modus ponens can be used to apply a rule to a fact. For example applying ``Cats are birds'' to ``Sam is a cat'' gives a new fact ``Sam is a bird''. To use bottom-up reasoning: \\
        - List the rules which do not contain specific named instances.\\
        - List all given facts with a specific named instance as a current facts. \\
        - Then for each rule do the following:\\
        1.) Examine each rule to see if its premise applies to any of the current facts. It's important to recognize that general rules can apply to specific facts. For example, the rule 'All cats are birds' applies to the fact `Sam is a cat,' even though the rule doesn't mention Sam specifically. This is because the rule involves a category ('cats') that matches a variable in the fact ('cat'). So, when checking for a match, look beyond direct mentions and consider whether the rule's general premise encompasses the specifics of the current facts. This careful matching is essential to correctly apply rules to facts. Then add the new concluded fact (in this example ``Sam is a bird'') to the list of current facts.\\
        2.) Check if the current facts contain the query or its negation. If they do, answer the query. If it does not, repeat this procedure for all rules again.
        Iterate through this process until a conclusion is reached. It may take three or four iterations to reach a conclusion that will answer the query.\\
        Statements: \{question \} \\
        Query:\{ query \}''}}
\end{quote}

\paragraph{\textbf{Top down}}
\begin{quote}
    \footnotesize{prompt = \textit{``Consider the following statements, which include rules and then facts, along with the given query. Use a top-down reasoning strategy to answer the query. Facts have a specific named instance. For example, `Sam is a cat' is a fact because an explicit name `Sam' is given. Rules establish a connection between two general classes without referring to a specific instance. For example, `Cats are birds' is a rule, not a fact, because it does not name a specific cat or bird. To use top-down reasoning, follow these steps:\\
    1. List the rules which do not contain specific named instances.\\
    2. List all given facts with a specific named instance as current facts.\\
    3. Then consider the query and repeat the following procedure:\\
    4. Check if the query or its negation is among the facts. If so, answer the query.\\
    5. If it is not, then search through the set of rules and check if the conclusion of any of the rules matches the query. If so, make the body of this rule the new query, paying special attention to maintaining the correct use of negation throughout the process. For example, if your query is `Does Sam not swim?' and you have a rule like `Birds swim', you would update the query to `Is Sam not a bird?'. Re-write the query with this new perspective and return to step 4.\\
    You will need to repeat the procedure multiple times until the query or its negation appear as facts. It is crucial to meticulously track all instances of 'not' or other negations when updating the query, as this directly affects the final answer.\\
    It may take three or four iterations to arrive at the final answer. If a particular line of reasoning comes to a dead end, it might help to start again with the original query and try applying different rules.\\
Statements: \{question\}\\
Query: \{query\}''}}
\end{quote}

\paragraph{\textbf{Magic Set}}
\begin{quote}
    \footnotesize{prompt = \textit{``Consider the following statements, which include both rules and facts, along with the given query. Approach the query with a top-down reasoning strategy, keeping in mind that facts are specific instances like `Sam is a cat,' and rules are general principles like `Cats are birds.' Execute the following steps:\\
1. List the general rules and specific named facts.\\
2. Create a set of subgoals starting with the query, aiming to gradually uncover the truth of the query through logical deduction.\\
3. Expand subgoals by systematically adding the premises of rules that directly relate to these subgoals. Continue this process until no new subgoals emerge.\\
4. Refine your set of rules by eliminating any that don't contribute to achieving the subgoals.\\
5. Carefully match the premises of the refined rules with the listed facts. Consider how general rules apply to specific instances and derive any new facts from this matching. Be particularly attentive to the implications of negations and complex statements.\\
6. Check if the current set of facts directly resolves the query or its negation. If a resolution is found, provide the answer.\\
7. If the answer is not yet clear, reapply the refined rules and facts iteratively, with each iteration aiming to reveal new insights or facts that help answer the query. In each iteration, aim to either refine your understanding or deduce a new fact that edges closer to resolving the query.\\
Continue this iterative process of application and review until a definitive conclusion is reached regarding the query. The process demands careful analysis and might require several iterations, so remain diligent and focused on how each piece of information contributes to your understanding of the query.\\
Statements: \{question\}\\
Query: \{query\}''}}
\end{quote}

\paragraph{\textbf{Small Model Prompt}}
\begin{quote}
\footnotesize{
You are a very skilled and concise logical translator who takes natural language and expresses it into formal language with great accuracy. Your task is to translate the given natural language sentences into a simplified logical format. Be a literal and precise translator.

Follow these strict rules:

1.  **General Rules:** A sentence is a general rule if it describes a relationship between categories (e.g., "All cats are mammals", "Sheep are animals"). For these, you MUST use the rule format exactly: "For all x, if x is a Category1, then x is a Category2."

2.  **Specific Facts:** A sentence is a specific fact if it describes a property of a specific individual with a proper name. Some example names are "Rex", "Fae" and "Sally". For these, you MUST use the fact format exactly: "ProperName is a Category."

3.  **Predicate Naming:** Combine multi-word concepts into a single, PascalCase predicate. For example "prime number" becomes "PrimeNumber".

4.  **Negation:** If a statement is negative, you MUST include "not" in the translation.

5.  **Punctuation:** End every logical statement with a period.

6.  **Mark the Query:** The query is the final sentence of the input. You MUST translate the query and enclose it in triple question marks, like this: `??? X is a Y. ???'.

Do not add any other explanation or text.

Here are some examples, notice that they follow the format exactly and include no explanation:
\{examples\}

I am alerting you to the most common mistakes here so you can avoid them. The most common mistake occurs when plural nouns are predicates. Some translators may write:
"Cows are Angry." instead of "For all x, if x is a Cow, then x is Angry."
A second example of a plural noun predicate mistake is:
"Felines are not Feisty." is written instead of "For all x, if x is a Feline, then x is not Feisty."
And another example of an error, this is incorrect: "Animals are Spiders.", the correct version is: "For all x, if x is an Animal, then x is a Spider.".

The second common error is converting a single word adjective predicate into a noun. For example, an incorrect translation was "Max is not a FruityThing." instead of  the correct "Max is not a Fruity".

The third common error is to use "it" instead of "x". For example: "For all x, if x is a Lepidopteran, then it is Sweet." is wrong. Instead it should be "For all x, if x is a Lepidopteran, then x is Sweet."

You are a very good translator, do not make the mistakes of previous translators.

Now please write the simplified logic for the natural language below.
Natural Language:
\{problem\_nl\}

Simplified Logic Format:}
\end{quote}

\paragraph{\textbf{Small Model Repair Prompt}}
\begin{quote}
\footnotesize{
You are a very skilled and concise logical translator who takes natural language and expresses it into formal language with great accuracy. Your task is to translate the given natural language sentences into a simplified logical format. Be a literal and precise translator.

Follow these strict rules:

1.  **General Rules:** A sentence is a general rule if it describes a relationship between categories (e.g., "All cats are mammals", "Sheep are animals"). For these, you MUST use the rule format exactly: "For all x, if x is a Category1, then x is a Category2."

2.  **Specific Facts:** A sentence is a specific fact if it describes a property of a specific individual with a proper name. Some example names are "Rex", "Fae" and "Sally". For these, you MUST use the fact format exactly: "ProperName is a Category."

3.  **Predicate Naming:** Combine multi-word concepts into a single, PascalCase predicate. For example "prime number" becomes "PrimeNumber".

4.  **Negation:** If a statement is negative, you MUST include "not" in the translation.

5.  **Punctuation:** End every logical statement with a period.

6.  **Mark the Query:** The query is the final sentence of the input. You MUST translate the query and enclose it in triple question marks, like this: `??? X is a Y. ???'.

Do not add any other explanation or text.

Here are some examples, notice that they follow the format exactly and include no explanation:
\{examples\}

Here is the original natural language:
\{problem\_nl\}

Your previous translation of this natural language text resulted in a logical inconsistency. Read the original natural language again and provide a correct version of the simplified logic. Your previous translation contained errors in some lines. They were probably of the form of like "Cows are not Carnivores" instead of the "For all x, if x is a Cow, then x is not a Carnivore." format for some lines. Each line should either start with a Proper Noun (a specifically named individual) or have forall in it. If you find any exceptions to this, please repair them by inserting the correct forall format. Remember to keep the query marked with `???'.

Corrected Simplified Logic Format:}
\end{quote}

\end{document}